\documentclass[11pt]{article}
\usepackage{amssymb}
\usepackage{newunicodechar}
\newunicodechar{ℝ}{\(\mathbb{R}\)}
\newunicodechar{ℂ}{\(\mathbb{C}\)}

\usepackage[preprint]{acl}

\usepackage{times}
\usepackage{latexsym}
\usepackage{amssymb}
\usepackage{amsmath}
\usepackage{hyperref}
\usepackage{url}
\usepackage{booktabs}
\usepackage{multirow}
\usepackage{booktabs,multirow,tabularx,array}
\usepackage{makecell}
\usepackage{tikz}
\newcolumntype{C}{>{\centering\arraybackslash}X}

\usepackage{booktabs,multirow,tabularx,array,makecell}

\usepackage{xcolor,booktabs}

\usepackage{array}
\newcolumntype{B}{>{\color{blue}}c}
\newcolumntype{R}{>{\color{red}}c}
\usepackage[most]{tcolorbox}
\tcbset{colback=green!10!white, colframe=green!50!black, boxrule=0.5pt, arc=3pt}
\usepackage{booktabs,multirow,makecell,xcolor}

\usepackage{tikz}
\usetikzlibrary{arrows.meta}
\tikzset{chatarr/.style={-{Stealth[length=2.4mm]}, semithick}}
\usepackage{tcolorbox}
\usepackage{amsmath}
\usepackage{adjustbox}

\newcommand{\filledcircle}[1]{\tikz[baseline=(char.base)]{%
    \node[shape=circle,fill=black,text=white,inner sep=0.5pt,minimum size=3.5mm] (char) {\tiny\bfseries#1};}}

\usepackage[T1]{fontenc}

\usepackage[utf8]{inputenc}

\usepackage{microtype}

\usepackage{inconsolata}

\usepackage{graphicx}

\newcommand{\methodname}[0]{\texttt{RULES}}

\title{Reliable Use of Lemmas via Eligibility Reasoning and Section‑Aware Reinforcement Learning}

\author{
 \textbf{Zhikun Xu\textsuperscript{1}\thanks{Work done during research internship at AMD.}},
 \textbf{Xiaodong Yu\textsuperscript{2}},
 \textbf{Ben Zhou\textsuperscript{1}},
 \textbf{Jiang Liu\textsuperscript{2}},
 \textbf{Jialian Wu\textsuperscript{2}},
\\
 \textbf{Ze Wang\textsuperscript{2}},
 \textbf{Ximeng Sun\textsuperscript{2}},
 \textbf{Hao Chen\textsuperscript{2}},
 \textbf{Zicheng Liu\textsuperscript{2}}
\\[1ex]
 \textsuperscript{1}Arizona State University,
 \textsuperscript{2}Advanced Micro Devices, Inc.
 \\
 \texttt{zhikunxu@asu.edu}
}

\begin{document}
\maketitle
\begin{abstract}
Recent large language models (LLMs) perform strongly on mathematical benchmarks yet often misapply lemmas, importing conclusions without validating assumptions. We formalize lemma‑judging as a structured prediction task: given a statement and a candidate lemma, the model must output a precondition check and a conclusion‑utility check, from which a usefulness decision is derived. We present \methodname{}, which encodes this specification via a two‑section output and trains with reinforcement learning plus section‑aware loss masking to assign penalty to the section responsible for errors. Training and evaluation draw on diverse natural‑language and formal proof corpora; robustness is assessed with a held‑out perturbation suite; and end‑to‑end evaluation spans competition‑style, perturbation‑aligned, and theorem-based problems across various LLMs. Results show consistent in‑domain gains over both a vanilla model and a single‑label RL baseline, larger improvements on applicability‑breaking perturbations, and parity or modest gains on end‑to‑end tasks; ablations indicate that the two‑section outputs and section‑aware reinforcement are both necessary for robustness.
\end{abstract}

\section{Introduction}

Recent large language models (LLMs) achieve strong scores on mathematical benchmarks and can often generate proof‑like text \cite{naturalprover,openai2024gpt4technicalreport,yang2024qwen25mathtechnicalreportmathematical,shao2024deepseekmathpushinglimitsmathematical,li2025one,guo2025mathematicalprooflitmustest}. Yet their behavior frequently reflects pattern imitation rather than principled reasoning: models tend to reproduce familiar solution paths, and they can import a lemma’s conclusion without validating its hypotheses. Such behavior exploits surface correlations in the training distribution and undermines trustworthiness in mathematical reasoning. 

Inspired by the human mathematician's needs \cite{frieder2024largelanguagemodelsmathematicians}, we cast lemma-judging as a structured prediction problem: given a statement-lemma pair, the model must output: \filledcircle{1} a precondition check (validating if the lemma's assumptions hold in the current context) and \filledcircle{2} a conclusion-utility check (assessing if the lemma is actually helpful for the proof goal), from which a useful decision is derived. Our hypothesis is that requiring models to externalize eligibility reasoning and aligning credit assignment per section will reduce reliance on spurious surface cues, improve robustness and safety to applicability-breaking perturbations, and transfer to end-to-end tasks. The intuition mirrors standard proof practice, i.e., check the hypothesis, then use the lemma, and explains why single-label supervision obscure the locus of failure. 

We propose a training framework, \methodname{} (Reliable Use of Lemmas via Eligibility Reasoning and Section-Aware Reinforcement Learning). The training is based on \textit{Group Relative Policy Optimization} (GRPO) \cite{shao2024deepseekmathpushinglimitsmathematical} with a section-aware loss masking mechanism, which assigns penalty to the precise section responsible for failure, rather than the entire rollout. This design aligns the supervision with the cognitive decomposition the task demands, encouraging models to learn the check-before-use discipline rather than overfitting to label-level shortcuts.

Empirical results over various models and diverse evaluation datasets have shown that \methodname{} improves in-domain lemma-judging over a vanilla model and a single-label RL baseline. The margin is larger on the held-out applicability-breaking perturbation suite with at most 23 points, indicating better gating of lemma use. On end-to-end benchmarks, average performance is broadly better or on par with the single-label RL baseline. Ablations show that both the two-section outputs and section-aware RL are necessary: entangling fine-grained supervision on checks or excluding training-time perturbations supervision substantially reduces the robustness of eligibility reasoning.

\section{Related Works}
Many works have explored pre-training on math corpora coupled with chain-of-thought prompting, which equips LLMs with domain-specific understanding and step-by-step reasoning~\cite{wang2025survey}. Reinforcement learning advances, such as GRPO, have offered significant gains on benchmarks like GSM8K and MATH~\cite{shao2024deepseekmathpushinglimitsmathematical}. However, brittleness to premise order and problem phrasing remains a challenge~\cite{chen2024premise, huang2025mathperturb}. Self-correction methods have also improved LLM mathematical reasoning by enabling self-refinement~\cite{xiong2025self, yan2025s}. Benchmarks such as the Open Proof Corpus~\cite{dekoninck2025open} and MathArena~\cite{balunovic2025matharena} reveal a gap between the accuracy of final answers and the validity of proofs. Knowledge-graph agents have been explored to improve success rates~\cite{li2025automating}, while LeanDojo~\cite{yang2023leandojo} and other works~\cite{mikula23magnushammer, tao2025learning} enhance premise selection. Recent innovations~\cite{mukherjeepremise, pan2025lemma, wangimproving} further align supervision with intermediate reasoning, improving error detection and reflective reasoning. Our work builds on existing insights and focuses on structured intermediate reasoning that aims to improve generalizable LLM reasoning on math. 



\section{Methodology: \methodname{}}
\methodname{} has two components: \filledcircle{1} a \textbf{two‑section lemma‑judging task} that requires explicit eligibility checks before use, and \filledcircle{2} \textbf{section‑aware reinforcement learning} that aligns credit assignment with these checks.

\subsection{Lemma-judging Task}
Given an input $x=(L, S)$ with lemma $L$ and statement $S$, the model produces a structured output with two sections $y=(r_{pre}, l_{pre})\oplus(r_{con}, l_{con})$ where $r_{pre}$ and $r_{con}$ are free-form rationales for the precondition and conclusion-utility checks, and the section (predicted) labels, i.e., $l_{pre}, l_{con}\in\{0,1\}$, are extracted judgments. The final usefulness decision is a deterministic aggregation of the precondition and conclusion checks $l_{use} = l_{pre} \land l_{con}\in\{0,1\}$. This task specification makes the precondition and conclusion checks explicit and auditable, and a minimal schema with section sentinels ensures unambiguous parsing, which is illustrated in Appendix~\ref{appendix:prompt} by prompts.

\subsection{Section-Aware Reinforcement Learning}
The training is adapted from GRPO. Specifically, for each input $x$, given a rollout as $y^i=y^i_{pre}\oplus y^i_{con}=(r^i_{pre}, l^i_{pre})\oplus(r^i_{con}, l^i_{con})$, if the precondition check is correct but the aggregated final label is incorrect, which implicitly means that the conclusion check must be wrong, we will only keep the conclusion check section, i.e., setting the loss mask of $(r^i_{con}, l^i_{con})$ to be 1 and others as 0. For the RL update, the gradient function in each group $G$ is illustrated as: $\nabla_\theta L_{section-aware} = -\frac{1}{|G|}\sum_{i=1}^{|G|}\frac{1}{|y^i|}\sum_{j=1}^{|y^i|}m^i_j\tilde{A_{j}^i}\nabla_{\theta}\log\pi_\theta(y^i_j|x, y^i_{<j})$, where $m^i_j$ is the j-th token mask for the i-th rollout and other notations in this equation follow the default GRPO setting such as estimated advantages $\tilde{A_{j}^i}$.
By this, we are penalizing the incorrect sections only and supervising models in more fine-grained signals on certain wrong tokens, enforcing "check-before-use" as a safety filter for theorem application.
Crucially, this objective-level modification operates independently of the underlying model architecture or scale. The computational overhead is restricted to negligible token-masking operations during the loss calculation, making \methodname{} compatible with any LLM backbones.

\begin{table*}[ht]
\centering
\scriptsize
\vspace{-7ex}
\begin{tabular}{@{}lcccccccccccc@{}}
\toprule
\multicolumn{2}{c}{\multirow{2}{*}{\textbf{Models \textbackslash Tests}}} &
  \multicolumn{4}{c}{\textbf{In-Domain Test}} &
  \textbf{Perturbation Test} &
  \multicolumn{6}{c}{\textbf{E2E Benchmark}} \\ \cmidrule(l){3-13} 
\multicolumn{2}{c}{} &
  \textbf{NP} &
  \textbf{NLPS} &
  \textbf{IS} &
  \textbf{IL} &
  \textbf{DT} &
  \textbf{PA} &
  \textbf{CM} &
  \textbf{IM} &
  \textbf{MP-hard} &
  \textbf{MP-simple} &
  \textbf{TQ-Math} \\ \cmidrule(r){1-2}
\multirow{3}{*}{DeepMath} &
  vanilla &
  60.8 &
  51.5 &
  51.6 &
  31.9 &
  24 &
  39.1 &
  45.7 &
  \textbf{76} &
  53 &
  72 &
  55.4 \\
 &
  GRPO &
  77.5 &
  77.2 &
  67.6 &
  65.8 &
  35.6 &
  37.9 &
  \textbf{63.1} &
  73 &
  54.5 &
  72.8 &
  56.1 \\
 &
  RULES &
  \textbf{88.5} &
  \textbf{86.2} &
  \textbf{75.8} &
  \textbf{89.5} &
  \textbf{47} &
  \textbf{39.5} &
  46 &
  71 &
  \textbf{56.3} &
  \textbf{72.8} &
  \textbf{57.7} \\ \midrule
\multirow{3}{*}{OLMO2} &
  vanilla &
  77.6 &
  69.1 &
  53.9 &
  64.9 &
  47.6 &
  \textbf{8.4} &
  \textbf{52.3} &
  \textbf{29} &
  8.6 &
  18.6 &
  27.1 \\
 &
  GRPO &
  \textbf{85.2} &
  \textbf{84.2} &
  72.2 &
  \textbf{83.6} &
  46.8 &
  7.8 &
  48.1 &
  23 &
  9.3 &
  21.9 &
  \textbf{30.3} \\
 &
  RULES &
  84.9 &
  83.4 &
  \textbf{76.6} &
  82.5 &
  \textbf{60.5} &
  6.7 &
  48.2 &
  24 &
  \textbf{10.4} &
  \textbf{23.3} &
  26.9 \\ \midrule
\multirow{3}{*}{Qwen2.5} &
  vanilla &
  63.9 &
  71.9 &
  55.5 &
  55.8 &
  71.9 &
  32.4 &
  59.5 &
  \textbf{55} &
  33.7 &
  59.5 &
  49.8 \\
 &
  GRPO &
  76.5 &
  83.7 &
  74.2 &
  80.4 &
  81.2 &
  \textbf{35.6} &
  60.5 &
  54 &
  33.7 &
  59.5 &
  50 \\
 &
  RULES &
  \textbf{88.2} &
  \textbf{85.8} &
  \textbf{81.3} &
  \textbf{92.7} &
  \textbf{87.8} &
  32 &
  \textbf{61.3} &
  53 &
  \textbf{35.1} &
  \textbf{62.4} &
  \textbf{50.5} \\ \midrule
\multirow{3}{*}{Llama-3.1} &
  vanilla &
  61.3 &
  75.1 &
  54.7 &
  52.3 &
  89.2 &
  11.5 &
  \textbf{56.5} &
  \textbf{22} &
  \textbf{13.3} &
  \textbf{38.7} &
  \textbf{33.5} \\
 &
  GRPO &
  88.6 &
  \textbf{86.5} &
  85.9 &
  89.8 &
  80.9 &
  \textbf{13.2} &
  50.7 &
  18 &
  13.3 &
  36.2 &
  32.6 \\
 &
  RULES &
  \textbf{93.2} &
  86.4 &
  \textbf{86.3} &
  \textbf{93.6} &
  \textbf{90.9} &
  12.8 &
  49.3 &
  16 &
  12.9 &
  36.9 &
  33 \\ \bottomrule
\end{tabular}
\caption{\textbf{Main Results}. For in-domain test, NP, NLPS, IS, and IL stands for NaturalProofs, NLPS, Premise Selection in Isabelle, IMO-Lemma derived test sets. DT stands for DeepTheorem-sampled perturbation test set. PA, CM, IM, MP, and TQ stand for Putnam-Axiom, CounterMATH, IneqMath, MATH-Perturb, and TheoremQA. MP-hard and MP-simple stand for hard and simple perturbations based on the paper. TQ-Math is the math subset of the TQ benchmark. The CounterMATH is using F1 due to imbalanced data distribution mentioned in \citet{li2025one} and the rest of data is using exact-match based accuracy.}
\label{tab:main}
\end{table*}

\begin{table*}[ht]
\centering
\scriptsize
\begin{tabular}{@{}lccccccccccc@{}}
\toprule
\multicolumn{1}{c}{\multirow{2}{*}{\textbf{Ablations}}} &
  \multicolumn{4}{c}{\textbf{In-Domain Test}} &
  \textbf{Perturbation Test} &
  \multicolumn{6}{c}{\textbf{E2E Benchmark}} \\ \cmidrule(l){2-12} 
\multicolumn{1}{c}{} &
  \textbf{NP} &
  \textbf{NLPS} &
  \textbf{IS} &
  \textbf{IL} &
  \textbf{DT} &
  \textbf{PA} &
  \textbf{CM} &
  \textbf{IM} &
  \textbf{MP-hard} &
  \textbf{MP-simple} &
  \textbf{TQ-Math} \\ \cmidrule(r){1-1}
vanilla &
  63.9 &
  71.9 &
  55.5 &
  55.8 &
  71.9 &
  \underline{32.4} &
  59.5 &
  \textbf{55} &
  33.7 &
  59.5 &
  49.8 \\
GRPO &
  76.5 &
  83.7 &
  74.2 &
  80.4 &
  \underline{81.2} &
  \textbf{35.6} &
  \underline{60.5} &
  \underline{54} &
  33.7 &
  59.5 &
  50 \\
two-section-onetime &
  87 &
  85.4 &
  \textbf{84.4} &
  92.7 &
  73.1 &
  \underline{32.4} &
  \underline{60.5} &
  49 &
  \underline{35.5} &
  \underline{60.9} &
  49.5 \\
w/o perturbation data &
  \textbf{93.2} &
  \textbf{87.4} &
  82.4 &
  \underline{94.4} &
  19.1 &
  32.2 &
  56.4 &
  49 &
  34.4 &
  59.1 &
  \underline{50.5} \\
de-noised perturbation data &
  86 &
  83.9 &
  \underline{83.6} &
  \textbf{95} &
  74.3 &
  34.1 &
  59.6 &
  \underline{54} &
  \textbf{35.8} &
  58.8 &
  \textbf{50.9} \\\midrule
RULES &
  \underline{88.2} &
  \underline{85.8} &
  81.3 &
  92.7 &
  \textbf{87.8} &
  32 &
  \textbf{61.3} &
  53 &
  35.1 &
  \textbf{62.4} &
  \underline{50.5} \\ \bottomrule
\end{tabular}
\caption{\textbf{Ablation Results.} \textit{two-section-onetime} refers to output two checks predictions at last. \textit{w/o perturbation data} and \textit{de-noised perturbation data} refers to removing perturbation data in training and filter the unqualified perturbation training data by the o3 model and human validation, respectively. The experiments are done with Qwen2.5-Math-7B-Instruct.}
\label{tab:ablation}
\end{table*}

\section{Experimental Settings}
\subsection{Models}
We have mainly experimented with four LLMs: \textbf{DeepMath-1.5B} \cite{he2025deepmath103klargescalechallengingdecontaminated}, \textbf{OLMo-2-1124-7B-Instruct} \cite{olmo20252olmo2furious}, \textbf{Qwen2.5-Math-7B-Instruct} \cite{yang2024qwen25mathtechnicalreportmathematical}, and \textbf{Llama-3.1-8B-Instruct} \cite{grattafiori2024llama3herdmodels}. For each model, we report three variants: \filledcircle{1} \textbf{Vanilla}: direct inference with our two-section schema. \filledcircle{2} \textbf{GRPO}: default GRPO optimization with only the final usefulness label (single-label objective) using the vanilla prompt. \filledcircle{3} \textbf{\methodname{}}: GRPO with section-aware masking by two-section schema.
\subsection{Training Settings}
The training data are from natural language and formal proof corpora, i.e., \textbf{NaturalProofs} \cite{welleck2021naturalproofs}, \textbf{NLPS} (Natural Language Premise Selection) \cite{ferreira-freitas-2020-natural}, \textbf{Premise Selection in Isabelle} \cite{mikula23magnushammer}, and \textbf{IMO-lemmas} \cite{liang2024decoupling}, pairing statement with candidate lemmas and binary labels showing the usefulness of lemmas. More training settings and designs are in Appendix~\ref{appendix:training}.

\paragraph{Perturbation Design.}As our goal is to separate surface similarity from true eligibility, we also add some perturbation data in training by making a lemma inapplicable while keeping the statement minimally changed and mathematically coherent. The reason of only providing labels for the precondition check when constructing perturbations is two fold: \filledcircle{1} Precondition satisfaction is directly falsifiable from the statement-lemma pair and independent of proof strategy, yielding reliable supervision with low ambiguity. \filledcircle{2} Conclusion utility is inherently strategy-dependent (e.g., proof by counterexample), making it costly and noisy to annotate at scale.

\paragraph{Reward Design.} For a fair comparison with baselines, the reward is rule-based and binary, i.e., 1 for correct and -1 for incorrect. Moreover, outputs that fail to parse under the two-section schema receive a format penalty of -2.

\subsection{Evaluation Settings}
We evaluate three axes that mirror our intended use: in-domain lemma-judging, robustness to applicability-breaking perturbations, and transfer to end-to-end mathematical problem solving. All results are calculated by sampling 7 outputs and majority-voting the extracted answers by self-consistency, which is noted as \textbf{SC@7}. More evaluation details are shown in Appendix~\ref{appendix:eval}.

\paragraph{In-domain Evaluation}
We use held-out splits drawn from the training data sources, which are named as \textbf{naturalproofs-test}, \textbf{nlps-test}, \textbf{isabelle-test}, and \textbf{IMO-lemma-test}.

\paragraph{Perturbation Test}
Similar to the goal and process of training perturbation data curation, we also evaluate on a out-of-domain perturbation suite constructed by minimally editing statements to violate lemma preconditions. But the raw data is sampled from DeepTheorem \cite{zhang2025deeptheoremadvancingllmreasoning}, which is noted as \textbf{deeptheorem-perturbation}. More details about perturbation data quality are in Appendix~\ref{appendix:data-quality}.

\paragraph{End-to-end Benchmarks}
Besides the task-related evaluations, we also assess transfer on diverse mathematical tasks, including competition-style problems (\textbf{Putnam-Axiom} \cite{gulati2024putnamaxiom}), inequality reasoning (\textbf{IneqMath} \cite{sheng2025solvinginequalityproofslarge}), counterexample-driven questions (\textbf{CounterMATH} \cite{li2025one}), perturbation-aligned suites (\textbf{MATH-Perturb} \cite{huang2025mathperturb}), and theorem-centric math QA (\textbf{TheoremQA} \cite{chen-etal-2023-theoremqa}).

\section{Analysis}
\subsection{Main Results}
\paragraph{In-domain lemma-judging.} Across all four base models, \methodname{} improves in-domain lemma-judging over both \textit{vanilla} and the single-label \textit{GRPO} baseline. Gains are consistent on most splits and especially pronounced on the IMO-derived test, with OLMO2 showing comparable performance to GRPO on three splits and clear improvements on the Isabelle split. These results indicate that enforcing explicit precondition and conclusion checks could help models improve understanding of lemma applicability.

\paragraph{Robustness to applicability-breaking perturbations.}On the out-of-domain perturbation suite, \methodname{} consistently outperforms \textit{vanilla} and \textit{GRPO} across all four models, with sizable margins for 23+ points at most. This pattern supports the intended effect of training-time perturbation data and section-aware masking: the model more reliably gates off familiar but inapplicable lemmas under minimal statement edits that invalidate the preconditions.


\paragraph{Mixed E2E transfer.} On downstream tasks, \methodname{} performs broadly on par with single-label \textit{GRPO}, yielding gains on perturbation-aligned subsets while showing occasional regressions on competition-style and inequality-based problems. We interpret these regressions as a \textbf{performance-rigorousness tradeoff}: whereas standard models may sometimes ``get lucky'' by hallucinating applicability to force a solution, \methodname{} enforces strict precondition checks. This discipline prioritizes precision and safety against invalid reasoning, naturally lowering recall on tasks where looser constraints might inadvertently permit a correct answer. Consequently, gains are most pronounced on perturbation-aligned suites where robustness against invalid inputs is paramount. Conversely, constructive tasks like CounterMATH require example synthesis, a capability distinct from eligibility gating. Thus, \methodname{} functions as a necessary robustness layer, ensuring models do not attempt proofs with fundamentally inapplicable premises.

\subsection{Ablations}
\paragraph{Two-time v.s. one-time prediction.}Replacing the sequential two-section protocol with a ``two-section-onetime'' variant (emitting both judgements only at the end) yields strong in-domain accuracy but notably weaker robustness on the perturbation suite. This indicates that the staged structure and section-aware alignment matters for learning to withhold lemmas when preconditions are not satisfied. 


\paragraph{Role of training-time perturbation data.} Removing perturbation data during training preserves in-domain scores but collapses robustness on the out-of-domain suite, confirming that curated negatives are essential to teach the precondition gate. However, we emphasize that this robustness is not merely a function of data scaling. The architecture contributes significantly: \methodname{} improves over the data-equivalent ``two-section-onetime'' baseline by 14.7 points on the perturbation test (Table \ref{tab:ablation}). This substantial margin confirms that while the data provides the necessary signals, the \textbf{Section-Aware RL mechanism} is indispensable for effectively leveraging them to prevent superficial pattern matching.

\paragraph{Effect of "de-noised" perturbation data.} Filtering low-quality perturbations leads to mixed outcome: modest improvements on some end-to-end metrics and in-domain tests, but still below \methodname{} on the perturbation suite. This evidence thus far does not support that "clean fine-grained labels" always improve overall math reasoning; rather, there appears to be a trade-off between strict cleaning and robustness coverage. 

\section{Conclusion}
We propose \methodname{}, aligning lemma use with an explicit check-before-use protocol and section-aware RL. \methodname{} yields consistent in-domain gains, large robustness improvements under precondition-breaking perturbations, and parity or small gains on end-to-end tasks. Ablations show the staged two-section output and section-aligned credit assignment are both required.

\newpage
\section*{Limitations}
\paragraph{Task scope and supervision.} Our study focuses on \textit{lemma-judging} rather than full proof generation. The supervision signals are tailored to the specific task, which may bias models toward unexpected behaviors.

\paragraph{Label coverage and noise.} In the perturbation data, only the precondition label is available while the conclusion-utility label is omitted to avoid strategy dependence. This asymmetric supervision can introduce bias and limit learning of usefulness beyond eligibility. In addition, rewards are binary with a format penalty, which simplifies optimization but may underutilize graded signals.

\paragraph{Model and data coverage.} Experiments cover four mid-sized models and a specific suite of math corpora (NaturalProofs, NLPS, Isabelle premise selection, and IMO-lemma). This leaves gaps across subfields of mathematics and automated theorem proving techniques, and languages other than English. While the eligibility reasoning mechanism is theoretically language-agnostic, the extension to multilingual settings or other formal proof assistants (e.g., Lean, Coq) remains a direction for future work.


\bibliography{custom}

\appendix
\section{Prompts}
\label{appendix:prompt}
\begin{tcolorbox}[title=\footnotesize Two-Section Prompt,top=1mm,bottom=1mm]
\scriptsize
Lemma: \{lemma\}\\
Statement: \{statement\}\\
\\
Please analyze step by step whether the lemma is useful for proving or disproving the given statement. You should perform two sequential checks:\\
\\
**Step 1: Precondition Check**\\
First, analyze whether the lemma's preconditions are satisfied in the given statement. After your reasoning, provide your answer within boxed\{\} as either boxed\{True\} (preconditions are satisfied) or boxed\{False\} (preconditions are not satisfied).\\
\\
**Step 2: Conclusion Check**\\
Next, analyze whether the lemma's conclusion is helpful for proving or disproving the given statement. After your reasoning, provide your answer within boxed\{\} as either boxed\{True\} (conclusion is helpful) or boxed\{False\} (conclusion is not helpful).\\
\\
\\
Please use the same section titles as above (i.e., **Step 1: Precondition Check** and **Step 2: Conclusion Check**) to start the two checks respectively and STRICTLY follow the following format:\\
- First, provide your reasoning for the precondition check\\
- Then output boxed\{True\} or boxed\{False\} for the precondition check\\
- Then provide your reasoning for the conclusion check\\
- Finally output boxed\{True\} or boxed\{False\} for the conclusion check
\end{tcolorbox}

\begin{tcolorbox}[title=\footnotesize Vanilla Prompt,top=1mm,bottom=1mm]
\scriptsize
Lemma: \{lemma\}\\
Statement: \{statement\}\\
\\
Please reason step by step about whether the above lemma is useful for proving or disproving its following statement, and then put your final answer (i.e., True or False) within \\boxed{}.
\end{tcolorbox}

\begin{tcolorbox}[title=\footnotesize Two-Section Onetime Prompt,top=1mm,bottom=1mm]
\scriptsize
Lemma: \{lemma\}\\
Statement: \{statement\}\\
\\
Please think step by step whether the lemma is useful for proving or disproving the given statement. Specifically, please 1) check whether the lemma's preconditions are satisfied in the given statement, and 2) check whether the lemma's conclusion is helpful for proving the given statement. Then put your final answers for precondition check and conclusion check, i.e., True or False for each check, within boxed\{\} (For example, boxed\{True, False\} means the lemma's preconditions are satisfied but its conclusion is not helpful for proving the given statement).
\end{tcolorbox}

\begin{tcolorbox}[title=\footnotesize Precondition Perturbation Prompt for o3 (DeepTheorem),top=1mm,bottom=1mm]
\scriptsize
Based on the given statement and its proof, could you help me perturb the statement with the minimal edits so that one certain precondition of some lemma used in the proof is no longer satisfied? If the perturbation is not possible, please output suitability: False. Otherwise, return suitability: True, and then extract the affected lemma in the proof, generate the perturbed statement, and the rationale of the perturbation.\\
\\
Critical Requirements:\\
1. MATHEMATICAL COHERENCE: The perturbed statement must be mathematically well-defined and meaningful. Do not create statements that are conceptually impossible or nonsensical (e.g., "subgroups of semigroups").\\
\\
2. GENUINE INAPPLICABILITY: The chosen lemma must be fundamentally unusable in ANY reasonable proof approach for the perturbed statement. Avoid superficial changes where:\\
   - The lemma applies to a subset/restriction of the new domain\\
   - The lemma can be used indirectly through standard correspondences\\
   - The statement can be reduced back to the lemma's domain\\
   \\
3. VERIFICATION TEST: Before finalizing, ask: "Could the chosen lemma still contribute meaningfully to proving or disproving the perturbed statement through any mathematical pathway?" If yes, the perturbation is ineffective.\\
\\
Rules:\\
- Copy the OriginalStatement verbatim except for that one span.\\
- Do NOT strengthen assumptions, introduce unrelated concepts/symbols, or make changes that render the statement trivial.\\
- Prefer the simplest change that creates a fundamental barrier to lemma applicability.\\
- The perturbation should target the lemma's core assumptions or domain in a way that cannot be circumvented.\\
\\
Ineffective Perturbation Patterns to Avoid:\\
- Domain changes where standard correspondences exist (e.g., ℝ exponential → ℂ exponential when differentiating w.r.t. real variables)\\
- Structure downgrades that still allow the target concept (e.g., group → semigroup but keeping "subgroup" terminology)\\
- Changes that only affect surface terminology but not mathematical substance\\
\\
Return ONLY this schema:\\
Suitability: \{bool\}\\
AffectedLemma: \{lemma-content\}\\
OriginalStatement: \{original statement\}\\
PerturbedStatement: \{perturbed statement\}\\
Rationale: \{rationale explaining why the lemma becomes fundamentally inapplicable, not just surface-level different\}\\
\\
Now Process:\\
\\
Statement: \{statement\}\\
Proof: \{proof\}
\end{tcolorbox}

\begin{tcolorbox}[title=\footnotesize Precondition Perturbation Validation Prompt for o3,top=1mm,bottom=1mm]
\scriptsize
You are evaluating the quality of a mathematical statement perturbation. Your task is to determine if the perturbation effectively breaks the chosen lemma's applicability.\\
\\
Given Perturbation Data:\\
Lemma: \{lemma\}\\
Original Statement: \{statement\}\\
Perturbed Statement: \{perturbed statement\}\\
Rationale: \{rationale\}\\
\\
Evaluation Criteria (ALL must pass):\\
\\
1. MATHEMATICAL COHERENCE CHECK:\\
   - Is the perturbed statement mathematically well-defined?\\
   - Are all mathematical concepts used consistently and meaningfully?\\
   - Do the mathematical objects and relations make sense together?\\
   \\
   Common failures:\\
   - Mixing incompatible concepts (e.g., "subgroups of semigroups")\\
   - Ill-defined operations or structures\\
   - Conceptual contradictions\\
\\
2. LEMMA INAPPLICABILITY VERIFICATION:\\
   Systematically check if the lemma could still be used via:\\
   \\
   a) DIRECT APPLICATION:\\
      - Can the lemma's hypotheses still be satisfied in the new context?\\
      \\
   b) RESTRICTION/SUBSET APPLICATION:\\
      - Does the lemma apply to relevant subsets of the new domain?\\
      - Can we restrict to cases where the lemma's conditions hold?\\
      \\
   c) CORRESPONDENCE/ISOMORPHISM:\\
      - Are there standard mathematical correspondences between the old and new domains?\\
      - Can we map the problem back to where the lemma applies?\\
      \\
   d) INDIRECT/FOUNDATIONAL USE:\\
      - Could the lemma be used in proving prerequisite results?\\
      - Is the lemma foundational to the theory needed for the perturbed statement?\\
      - Could it be applied to finite quotients, local analysis, etc.?\\
      \\
   e) ANALOGOUS RESULTS:\\
      - Are there direct analogs of the lemma in the new domain?\\
      - Would the original lemma be a key step in proving such analogs?\\
\\
3. FAILURE PATTERN DETECTION:\\
   Check for these known ineffective patterns:\\
   \\
   - SUPERFICIAL DOMAIN CHANGES: Changes that look different but preserve mathematical substance\\
   - TERMINOLOGY SWAPS: Changing labels without affecting underlying mathematics\\
   - TRIVIAL GENERALIZATIONS: Moving to broader categories where restrictions still apply\\
   - INCOMPLETE DOMAIN SHIFTS: Changing some but not all relevant aspects of the domain\\
\\
4. PROOF PATHWAY ANALYSIS:\\
   Consider: "In any reasonable attempt to prove or disprove the perturbed statement, could the given lemma contribute meaningfully to the argument?"\\
   \\
   If YES to any pathway → INEFFECTIVE perturbation\\
   If NO to all pathways → EFFECTIVE perturbation\\

Return your evaluation in this format:\\
\\
EFFECTIVENESS: [BOOLEAN]\\
RATIONALE: [Explanation of why the perturbation is effective or ineffective]\\
\\
Now evaluate the given perturbation:
\end{tcolorbox}

\section{More Training Details}
\label{appendix:training}
\begin{figure*}[!t]
\begin{center}
    \adjustbox{center}{\includegraphics[width=\textwidth]{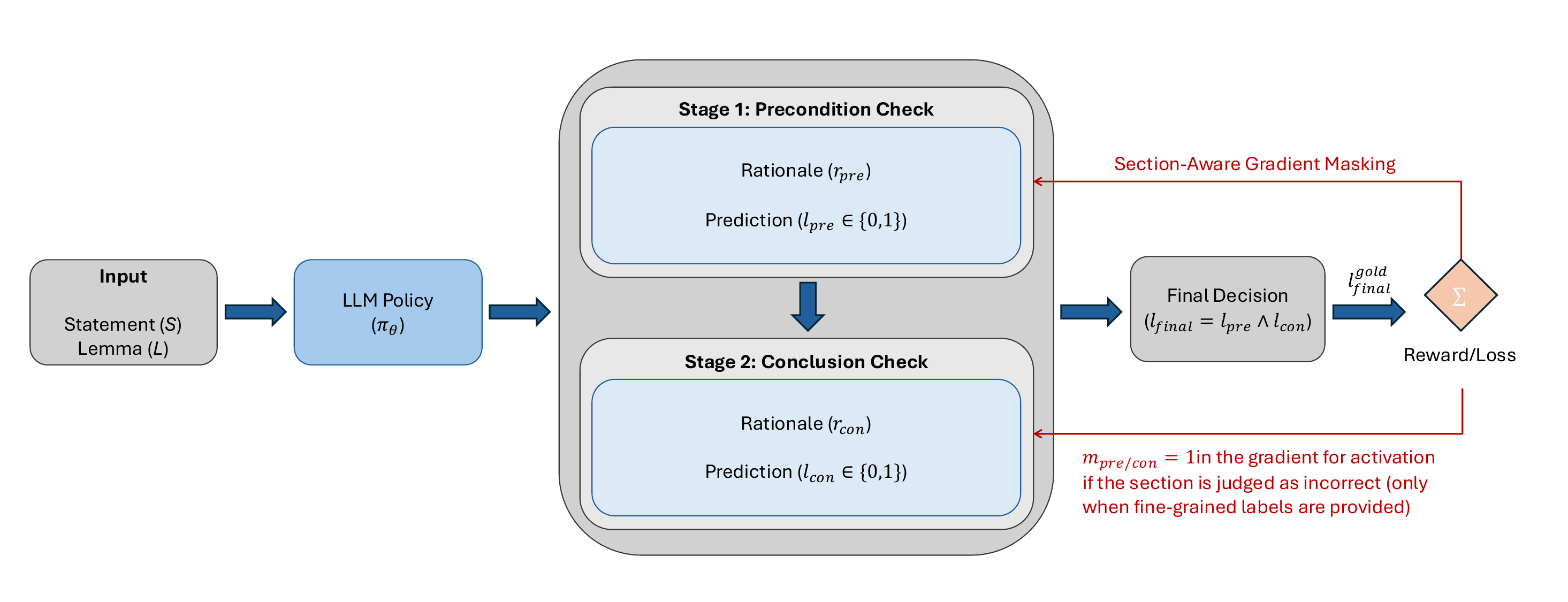}}
    \caption{\label{fig:pipeline}Overview of \methodname{} Training Framework. As mentioned in Section 3.2, if the fine-grained label (precondition/conclusion check) is provided, we could know which section of tokens are to blame with negative rewarding.}
\end{center}
\end{figure*}
Our training experiments are mainly implemented on AMD MI250 and Nvidia H200 GPU servers.  The training batch size is 128 and the number of rollout is set to 8. The response length is 8000 and 2048 for reasoning and non-reasoning models respectively. Training checkpoints are used when the percentage of activated section aware loss masking instances of a single batch is close to zero. Our training framework is based on \textit{verl} \cite{verl}. The training data distribution is NaturalProofs: 4744, NLPS: 2858, Premise Selection in Isabelle: 4484, and IMO-lemma: 986.

\paragraph{Details of Perturbation Design.}Concretely, we target a specific hypothesis of the lemma used in a proof and apply a minimal edit to the statement so that the hypothesis fails ($l^{gold}_{pre}=0$), without introducing unrelated concepts or trivializing the claim. This is derived from NaturalProofs and processed by o3 model\footnote{\url{https://platform.openai.com/docs/models/o3}}. The perturbation data is the only part with the gold precondition label for section-aware masking.

\section{More Evaluation Details}
\label{appendix:eval}
Our evaluation framework is based on \textit{vllm} \cite{kwon2023efficient}. The max token length is set to 10k to avoid early truncation and temperature is set to 0.7 for SC@7 across all models. The in-domain and perturbation evaluation data distributions are \textbf{naturalproofs-test} (1002), \textbf{nlps-test} (676), \textbf{isabelle-test} (256), \textbf{IMO-lemma-test} (342), and deeptheorem-perturbation (839)

\section{Perturbation Data Quality}
\label{appendix:data-quality}
To verify the quality of our perturbation suite (where only the precondition label is flipped), we conducted a multi-stage validation. For the training data, we have used O3 model and GPT-OSS-120B to cross-validate them, where both models reached consensus on 84.8\% of training data. We only use these 84.8\% data as our final perturbation training. Moreover, we authors have also sampled 100 datapoints from it for human verification, and the pass rate is 90\%. For perturbation evaluation data, we authors have sampled 100 datapoints after o3-based and human filtering to verify the quality of the final data. Our evaluation guideline follows exactly the principles in the prompt: (i) mathematical coherence (ii) genuine inapplicability (iii) verification for similar pathways by equivalent lemmas. The pass rate for the sampled evaluation data is 95\%.

\section{Case Study: Preventing Lemma Misapplication}
To illustrate how \methodname{} prevents the hallucinations of lemma applicability, we present a case from the Riemannian Geometry domain where a standard vanilla model fails.

\paragraph{Scenario.}
\textbf{Statement:} Let $G=SU(n)$ and let $g$ be a \textit{left-invariant} metric. Show that any symmetric 2-tensor $h$ satisfying the linearized Einstein equation... must be trivial. \\
\textbf{Candidate Lemma:} "Since $g$ is \textit{bi-invariant}, it is an Einstein metric... that $Ric_g=\lambda g$ for some $\lambda>0$."

\paragraph{Analysis.}
The lemma requires the metric to be \textit{bi-invariant}. However, the problem statement only guarantees that $g$ is \textit{left-invariant}. In differential geometry, left-invariance is a weaker condition than bi-invariance; thus, the lemma's preconditions are not met.

\paragraph{Model Behavior.}
\begin{itemize}
    \item \textbf{Baseline (Vanilla):} The model ignores the subtle distinction between "left-invariant" and "bi-invariant," hallucinates that the condition is met, outputs \texttt{True}, and attempts to use the lemma.
    \item \textbf{RULES:} The model correctly identifies the mismatch in the \textit{Precondition Check} section:
    \textit{"Rationale: The lemma requires bi-invariance, but the statement only assumes left-invariance..."}
    It outputs \boxed{False} for the precondition, effectively gating off the invalid path.
\end{itemize}

\section{Use of LLM}
We have only used LLM for language polishing purposes in the paper writing. We do not use LLM for idea formalization, or to an extent that it could be regarded as a contributor.

\end{document}